%% file: main.tex
\documentclass[letterpaper]{article} 
\usepackage{aaai25}  
\usepackage{times}  
\usepackage{helvet}  
\usepackage{courier}  
\usepackage[hyphens]{url}  
\usepackage{graphicx} 
\usepackage{natbib}  
\usepackage{caption} 
\usepackage{algorithm}
\usepackage{algorithmic}

\usepackage{newfloat}
\usepackage{listings}
\DeclareCaptionStyle{ruled}{labelfont=normalfont,labelsep=colon,strut=off} 
\floatstyle{ruled}
\newfloat{listing}{tb}{lst}{}
\floatname{listing}{Listing}
\usepackage[utf8]{inputenc}
\usepackage[T1]{fontenc}
\usepackage{booktabs}
\usepackage{amsfonts}
\usepackage{nicefrac}
\usepackage{microtype}
\usepackage{xcolor}
\usepackage{algorithm}
\usepackage{algorithmic}
\usepackage{subcaption}
\usepackage{ifsym}
\usepackage{diagbox}
\usepackage{float}
\usepackage{multirow}
\usepackage{epsfig}
\usepackage{marvosym}
\usepackage{amsmath}
\usepackage{amssymb}
\usepackage{mathtools}
\usepackage{amsthm}
\usepackage{wasysym}
\usepackage{enumitem}
\usepackage{tcolorbox}
\usepackage{makecell}
\usepackage{colortbl}
\usepackage{tikz}

\definecolor{mygreen}{HTML}{3cb44b}
\definecolor{mygreyr}{rgb}{0.99,0.985,0.98}
\definecolor{mygreyr1}{rgb}{0.98,0.975,0.97}
\definecolor{mygreyr2}{rgb}{0.97,0.965,0.96}
\definecolor{mygreyr3}{rgb}{0.96,0.955,0.95}
\definecolor{mygreyr4}{rgb}{0.955,0.95,0.945}
\definecolor{myred}{rgb}{0.995,0.9796,0.958}
\definecolor{myred1}{rgb}{0.992,0.9576,0.932}
\definecolor{myred2}{rgb}{0.991,0.9416,0.917}
\definecolor{myred3}{rgb}{0.990,0.9256,0.902}
\definecolor{myred4}{rgb}{0.990,0.915,0.89}

\title{Capability Instruction Tuning: A New Paradigm for Dynamic LLM Routing}
\author{
      Yi-Kai Zhang, \ \textbf{De-Chuan Zhan}, \  \textbf{Han-Jia Ye}\thanks{Corresponding author.}
}
\affiliations{
  School of Artificial Intelligence, Nanjing University\\
  National Key Laboratory for Novel Software Technology, Nanjing University\\
  \{zhangyk, zhandc, yehj\}@lamda.nju.edu.cn
}

\usepackage{bibentry}

\begin{document}

\maketitle

\begin{abstract}
Large Language Models (LLMs) have demonstrated human-like instruction-following abilities, particularly those exceeding 100 billion parameters.
The combined capability of some smaller, resource-friendly LLMs can address most of the instructions that larger LLMs excel at.
In this work, we explore how to route the best-performing LLM for each instruction to achieve better overall performance.
We develop a new paradigm, constructing \textit{capability instructions} with model capability representation, user instruction, and performance inquiry prompts to assess the performance.
To learn from capability instructions, we introduce a new end-to-end framework called \underline{Model} \underline{S}election with \underline{A}ptitude \underline{T}est (\textsc{Model-SAT}), which generates positive and negative samples based on what different models perform well or struggle with. \textsc{Model-SAT} uses a model capability encoder that extends its model representation to a lightweight LLM.
Our experiments show that \textsc{Model-SAT} understands the performance dimensions of candidate models and provides the probabilities of their capability to handle various instructions.
Additionally, during deployment, a new model can quickly infer its aptitude test results across 50 tasks, each with 20 shots. \textsc{Model-SAT} performs state-of-the-art model routing without candidate inference and in real-world new model-released scenarios. The code
is available at \url{https://github.com/Now-Join-Us/CIT-LLM-Routing}.

\end{abstract}

\input{sections/introduction}

\input{sections/preliminary_method}

\input{sections/experiments}

\section*{Acknowledgments}
This work is partially supported by NSFC (62376118, 62250069), Key Program of Jiangsu Science Foundation (BK20243012), Collaborative Innovation Center of Novel Software Technology and Industrialization.

\bibliography{aaai25}
\end{document}

%% file: sections/introduction.tex
\section{Introduction}
Large Language Models (LLMs)~\cite{openai2020chatgpt,du2022glm,touvron2023llama,vicuna2023,jiang2023mistral} rapidly evolve, demonstrating near-human general capabilities, especially in understanding, reasoning, and creative tasks related to instruction-response scenarios. Recent advancements have even enabled these LLMs to be trained in multilingual~\cite{yang2024qwen2technicalreport,dubey2024llama3herdmodels}, multidomain~\cite{DBLP:journals/corr/abs-2406-04614,yang2024qwen2technicalreport}, and multimodal~\cite{chen2015microsoft,chen2023sharegpt4v,reid2024gemini1_5} environments, allowing them to tackle complex instructions such as ``What is the relationship between Fourier series and Hilbert space?'' or to interpret images by identifying, ``What are the basis vectors of the Hilbert space?''

The rise of LLMs and their extensions has incredibly energized community applications. However, achieving more comprehensive capabilities often requires LLMs of a larger scale. According to the Open LLM Leaderboard~\cite{myrzakhan2024openllmleaderboard}, 60\% of the top 50 LLMs have around 70 billion (B) parameters or more, with only three LLMs under 10B. Additionally, some closed-source LLMs consistently dominate performance rankings over extended periods.
Consequently, optimizing LLM applications often hinges on substantial computational resources or costly token purchases.
A natural idea arises: Can we utilize multiple smaller LLMs, which are more resource-friendly and have below one-tenth of the parameters of their larger counterparts, to achieve performance comparable to gigantic LLMs while maintaining low inference costs?

\begin{figure}[t]
    \centering
    \vspace{-10pt}
    \includegraphics[width=0.42\textwidth]{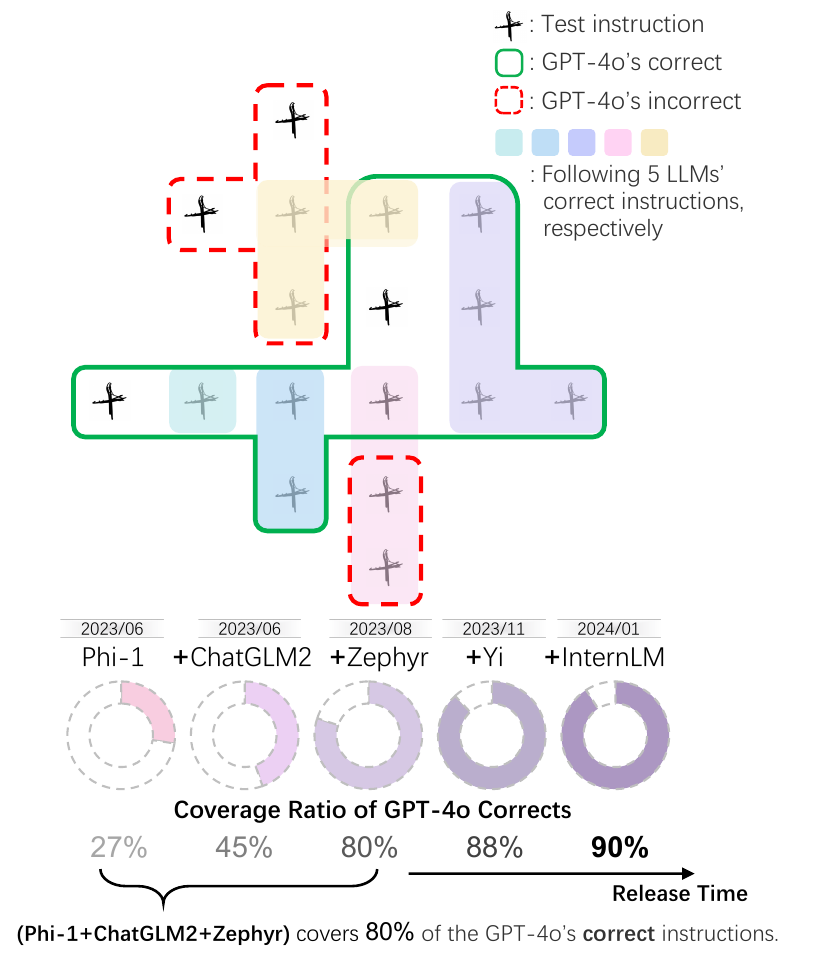}
    \caption{\textbf{Illustration of Coverage Observation}: The combined capabilities of the earlier-released model zoo effectively address most of the instructions that GPT-4o excels at. The union of samples managed accurately by Phi-1, ChatGLM2, and Zephyr covers 80\% of GPT-4o's correct instructions. The smaller-scale model zoo can enhance overall performance by selecting a suitable model for each instruction.}
    \label{fig:motivation}
    \vspace{-13pt}
\end{figure}

In the experiments, we find that the combined capability of some smaller-scale LLMs, despite their lower overall performance, can address most of the instructions that larger LLMs excel at. As shown in Figure~\ref{fig:motivation}, on the Massive Multitask Language Understanding (MMLU)~\cite{mmlu} benchmark, the Phi-1 LLM with 1.3B performs nearly 50\% worse than GPT-4o. However, it exhibits similar effectiveness to GPT-4o in the \texttt{high} \texttt{school} \texttt{mathematics} category.
Moreover, we create an early-access LLM zoo that includes Phi-1~\cite{DBLP:journals/corr/abs-2306-11644} and four 7B LLMs, which were released a year earlier than GPT-4o and exhibit an approximately 30\% performance gap compared to GPT-4o.
However, the combined accurate responses from this zoo cover 90\% of which GPT-4o handles correctly and address nearly 80\% with which GPT-4o struggles. By strategically assigning instructions to the suitable LLM in the zoo, there is potential to exceed GPT-4o's performance by 15\%.
From this phenomenon, the model routing for each instruction enhances performance with seamless LLM transitions and minimal inference costs, all without user awareness.

The key to the proposed instruction-level model routing is to efficiently identify the optimal model from a vast pool of options, without prior access to the potential candidates' inference outputs~\cite{DBLP:journals/corr/abs-2311-06720,bge_embedding} or the target task's ground truth~\cite{you2022ranking,DBLP:conf/cvpr/PandyAUFM22}.
In this paper, we introduce \textsc{Model-SAT}: \textbf{Model} \textbf{S}election with \textbf{A}ptitude \textbf{T}est. Our approach leverages 50 core 20-shot tasks, where the model test result represents model capability. By learning the generalization relationships between the capability representations of the candidate models and the instructions to be assigned, we can select the most suitable model across various repositories and target instructions.

Driven by the model capability representation, the \textsc{Model-SAT} framework establishes a novel paradigm, denoted as capability instruction tuning. Capability instructions consist of a capability representation, a user instruction, and a prompt to probe whether the model can perform that instruction. Using extensive historical performance data, capability instruction tuning learns an implicit relationship between core capability representations and unseen instructions. Moreover, it delves deeper into understanding the mapping between the capabilities' performance and the instructions' semantic distribution.
This intuition comes from the observation that individuals who perform well in the mathematical sections of the college admission SAT in the United States often pursue careers that involve logical reasoning. Capability instruction tuning aims to equip the model with a lightweight standardized guide to assess its effectiveness in handling future instructions.

Specifically, we combine model capability representation with positive and negative training instructions regarding current model performance, yielding statements like, ``The model achieves accuracy 85\% on the task of 'Mathematics, Geometry, ...'. Instruction: ..., Predict whether the model can handle ...''.
To align the performance distribution inherent in model representation to the instruction semantic, we are the first to incorporate a capability encoder and extend the input of a lightweight LLM to include capability representation. The end-to-end \textsc{Model-SAT} functions as a model router that outputs the probabilities indicating which models will likely excel at specific instructions.

Additionally, we establish several comprehensive benchmarks for model routing of LLMs and their extensions. Our benchmarks cover a range of model zoos, such as (\textbf{1}) smaller-scale, weaker ones, (\textbf{2}) mixed-scale options, and (\textbf{3}) high-performance larger-scale LLMs. Furthermore, we expand the model routing to include multimodal LLM-instruction settings.
\textsc{Model-SAT} achieve significantly improved overall performance across model zoos without incurring any inference overhead, comparable to the performance levels of larger-scale LLMs.
Notably, the capability instruction tuning maintains the model representation generalization to unseen data. The new LLM can quickly develop effective model representations after just a few inferences (only on 50 x 20-shot tasks). In light of practical routing scenarios with the emergence of new-version LLMs, we establish 60 incremental routing scenarios that impose higher routing speed and overhead requirements. Throughout these settings, \textsc{Model-SAT} consistently demonstrates superior performance.

In summary, our contributions are:
\begin{itemize}[noitemsep,topsep=0pt,leftmargin=*]
\item \textbf{A novel paradigm: capability instruction tuning}, where model representation with efficient aptitude tests and instructions create capability instructions for high-performance-driven instruction-level model routing.
\item \textbf{\textsc{Model-SAT} framework}, features a model capability encoder and a lightweight LLM to end-to-end learn the router via various model capability representations.
\item \textbf{Comprehensive model routing benchmarks for LLMs and their extensions}, covering five LLM zoo setups with multimodal scenarios, as well as simulating 60 incremental-released model routers to ensure quick adaptation to unseen data and new LLMs.
\item \textbf{An open-source, deployable model routing toolkit} that applies model routing techniques to any model zoo, enhancing performance while remaining unaware of users with acceptable routing delays.
\end{itemize}

%% file: sections/preliminary_method.tex
\section{Preliminary}

We begin by discussing the key elements and the pipeline of model selection, followed by the evolution of related works.

\begin{figure*}[t]
    \centering
    \vspace{-10pt}
    \includegraphics[width=0.98\textwidth]{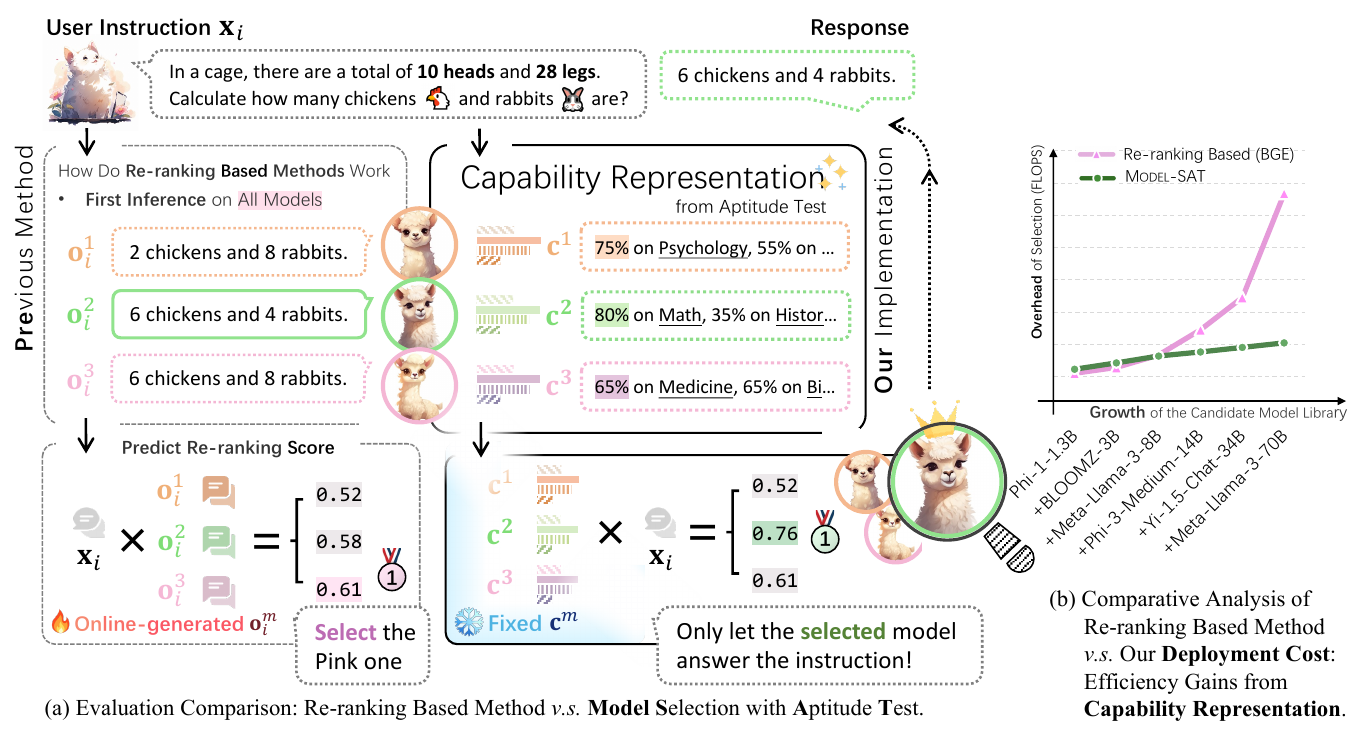}
    \caption{\textbf{Illustration of Model Routing with Capability Instructions: A Comparison with Re-ranking Based Methods.} The goal of model router is to select the optimal model for a given user instruction without access to ground truth and enhance overall performance. Previous re-ranking methods require inference for each candidate. \textsc{Model-SAT} employs a lightweight aptitude test to create capability representations. It learns the intrinsic relationship between model representations and the instructions to be assigned, significantly speeding up model routing and streamlining deployment.}
    \label{fig:related_works}
    \vspace{-10pt}
\end{figure*}

\subsection{Instruction, Output, and Answer}

Consider a test instruction dataset $\mathcal{D}_{\text{test}} = \left\{ \left(\mathbf{x}_i, \mathbf{a}_i \right)\right\}_{i=1}^{N}$ with $N$ labeled samples. The $\mathbf{x}_i$ and $\mathbf{a}_i$ represent the instruction and its corresponding answer, respectively.
Given an LLM or its extension, represented as $f$, the output generated for instruction $\mathbf{x}_i$ is denoted as $\mathbf{o}_i$, \textit{i.e.}, $f(\mathbf{x}_i) = \mathbf{o}_i$.
There are no restrictions on the language, domain, or modality of $\mathbf{x}_i$; In this paper, we focus on decoder-only text generation models, which means that $\mathbf{a}_i$ is typically presented in text form.
For the model $f$ to excel at instruction $\mathbf{x}_i$, it is equivalent to obtaining a high score on the evaluation $\operatorname{eval}\left( \mathbf{o}_i, \mathbf{a}_i \right)$.

\subsection{Pipeline of Model Routing}

Consider a candidate model zoo composed of many trained LLMs, $\mathcal{M} = \left\{f^m\right\}_{m=1}^{M}$. Model routing involves selecting a model from the zoo for each instruction $\mathbf{x}_i$ in the test dataset $\mathcal{D}_{\text{test}}$.
Specifically, the sequence of selected models is formalized as $\boldsymbol{f} = (f_1, f_2, \dots, f_N)$, where $f_i \in \mathcal{M}$.
We define the optimal model $\hat{f}$ for instruction $\mathbf{x}_i$ as the model that maximizes the score: $\operatorname{eval}\left(\hat{f}(\mathbf{x}_i), \mathbf{a}_i \right)$.
The objective of the instruction-level model routing is:
\begin{equation}
    \hat{\boldsymbol{f}} = \left( \underset{ {f}^m \in \mathcal{M} }{\arg \min } \; \ell \left({f}^m \left( \mathbf{x}_i \right),\; \mathbf{a}_i \right) \right)_{i = 1}^{N}\;,\label{eq:objective}
\end{equation}
where $\ell \left( \cdot \right)$ represents the loss function associated with the metrics between $\mathbf{o}^m_i = f^m(\mathbf{x}_i)$ and the ground truth $\mathbf{a}_i$.
The model routing bottleneck arises from the number of instructions on which no model in the zoo performs well.

\subsection{Revisit from Requirement, Target, and Key Inputs}

\textbf{Routing target} of \textit{parameter initialization} or \textit{models with zero-shot capabilities}: Early model router~\cite{tran2019transferability, nguyen2020leep,DBLP:conf/cvpr/TanLH21,DBLP:conf/eccv/DingCLCS22} efforts primarily focus on identifying a good training initialization that facilitates fine-tuning downstream tasks to achieve optimal performance. In this context, candidate models likely required additional training to adapt to the target task.
Recently, guided by scaling laws, foundational models like LLMs have experienced remarkable advancements in their zero-shot capabilities~\cite{touvron2023llama2,wei2022finetuned,team2023gemini}.
Extended models have demonstrated considerable potential in multilingual, multi-domain, and multimodal applications. For instance, Llama 3.1~\cite{dubey2024llama3herdmodels} serves as a multilingual agent, Qwen2-Math~\cite{yang2024qwen2technicalreport} tackles several Olympiad-level problems, and GPT-4o~\cite{openai2023gpt4} processes information from multiple sources.

\begin{figure*}[t]
    \centering
    \vspace{-10pt}
    \includegraphics[width=0.98\textwidth]{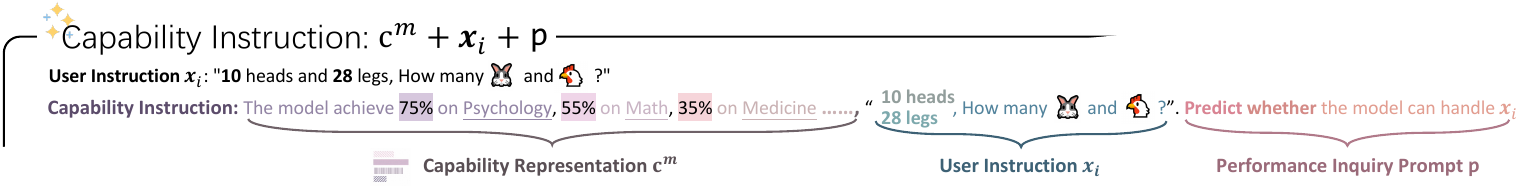}
    \caption{\textbf{One example of a Capability Instruction.} It is an instruction for model routing that inquires whether a model can handle a specific user instruction. It comprises three components: the capability representation $\texttt{c}^m$ based on the streamlined aptitude test, the user instruction $\mathbf{x}_i$ to be assigned, and a performance inquiry prompt $\texttt{p}$. This instruction is inputted into the \textsc{Model-SAT} Capability LLM, which outputs the probability that the model can perform the user instruction well.}
    \label{fig:capability_instruction}
    \vspace{-5pt}
\end{figure*}

\begin{figure}[t]
    \vspace{-10pt}
    \centering
    \includegraphics[width=0.48\textwidth]{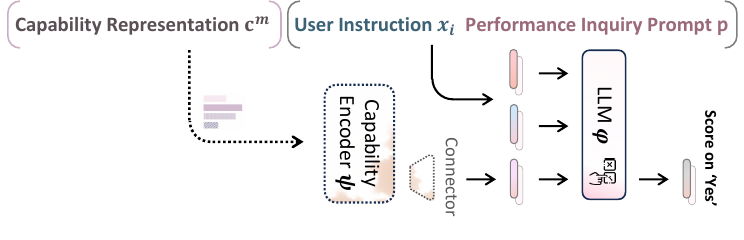}
    \caption{The Architecture of \textsc{Model-SAT}.}
    \label{fig:architecture}
    \vspace{-10pt}
\end{figure}

\textbf{Routing requirements} with \textit{target instruction annotation}, \textit{backpropagation delay}, or \textit{candidate output}: Some works~\cite{bao2019information,li2021ranking,DBLP:conf/icml/YouLWL21,deshpande2021linearized,DBLP:conf/cvpr/PandyAUFM22} design the proxy metric of transferability, which approximates the lower bound of fine-tuned performance. These works often rely on certain source clues, labeled instructions, or backpropagation steps to assess the transferability from the source pre-trained model to the target dataset.
Additionally, some re-ranking-based works~\cite{DBLP:journals/corr/abs-2311-06720,bge_embedding,zhang2024mgte} train an extra model to learn the contrastive relationships between the instruction and the candidate inference outputs $\left\{ \mathbf{o}^m_i \right\}_{m=1}^{M}$, routing the optimal one linked to model $f^m$.
However, obtaining all inferences may introduce significant delays when the number of models $M$ in the repository becomes excessively large~\cite{shnitzer2023large,lu2023routing,hu2024routerbench}.
Our \textsc{Model-SAT} aims to route models without annotation or inference requirements, considering candidates as black boxes.
A central feature is constructing model representations for each model and learning the adjusted relationship between it and the target instructions.

\textbf{Key input} -- \textit{model representation} for model routing:
When routing a model for instruction, the router requires the key representation that captures the model's characteristics. We followed the concept of learnware~\cite{DBLP:journals/fcsc/Zhou16a}, leveraging a small amount of model-proficient data to construct shared specifications~\cite{DBLP:journals/chinaf/ZhouT24,DBLP:conf/kdd/TanLBTZLXZYZ24}.
Other relevant methods leverage forward behavior or results on target as model representation, which inevitably introduces inference delays.
Recently, some approaches~\cite{lu2024blending,srivatsa2024harnessing,ding2024hybrid,DBLP:journals/corr/abs-2410-03834} have started to utilize learnable parameters as model representations.
For instance, some introduce a surrogate scorer as the corresponding model representation, learning the mapping from the task to the accuracy of candidate model outputs.
Model Spider~\cite{DBLP:conf/nips/ZhangHDZY23} takes this concept by encoding the model representation into a learnable vector, which acts as the input token for a Transformer-based router. However, learnable representation face challenges when new models are introduced, as they require extensive historical performance for costly pre-training of the router.
Our solution uses text-only descriptions of capabilities. New models can create representations by inferring 50 quick tasks, each with 20 shots.

%% file: sections/experiments.tex
\section{{\normalsize \textsc{Model-SAT}}: Model Routing with Aptitude Test}

In this section, we start by building the model representation and progress to the details of \textsc{Model-SAT}, training data, and optimization process. Finally, we outline an efficient deployment framework for model routing.

\subsection{Capability Instructions}
\label{sub_sec:capability_instructions}
The \textit{capability instruction} mainly comprises the capability representation of the candidate model $f^m$, user instruction $\mathbf{x}_i$, and performance inquiry prompt.
Specifically, the model's capability representation is formed from 50 distinct tasks across various categories from the MMLU dataset, with each task being 20-shot. We provide a concise description of five keywords for each task.
Next, we evaluate the candidate models across these 50 tasks and describe the results in natural language, \textit{i.e.}, model representation.
Furthermore, the advantage of representing in natural language is that it helps to include extra expert knowledge, such as mentioning \textit{which languages a model supports}.
The easy-to-obtain representations serve as an aptitude test for the models, indicating their potential capabilities across various dimensions.

To assess how well the candidate model can follow a single or a set of instructions $\mathbf{x}_i$, we introduce the training instructions that were executed correctly versus those incorrectly. These will be paired with the performance inquiry prompt $\texttt{p}$ to form the \textit{capability instruction}, denoted as $\mathbf{z}_i$, which drives the router to predict adaptation scores.
As illustrated in~Figure \ref{fig:capability_instruction}, it combines the capability representation $\texttt{c}^m$ for candidate $m$, the instruction $\mathbf{x}_i$, and a inquiry prompt $\texttt{p}$.

\noindent \textit{Core task sampling}: We sample instructions of core tasks with the highest distinguishability, avoiding those where most models perform correctly or incorrectly. In the model zoo of training, samples for which half of the models make mistakes while the other half are correct carry greater weight.

\subsection{Architecture}

\textbf{Motivation}: Although LLMs demonstrate strong instruction-following abilities, a gap exists between performance and the semantic distribution in \textit{capability instructions}, particularly in understanding combinations of performance dimensions. For example, if a candidate model achieves 80\% in mathematics and 95\% in legal principles, the model may possess legal reasoning skills.
To address this, we propose extending a capability encoder E5-Large \textasciitilde0.5B ($\boldsymbol{\psi}$) before a Phi-3-Mini 3.8B LLM ($\boldsymbol{\varphi}$) to align the candidate performance with the instructions. The architecture is illustrated in Figure~\ref{fig:architecture}.

\textbf{Structure}: The capability instruction comprises the capability representation $\texttt{c}^m$ for model $m$, the instruction $\mathbf{x}_i$, and the query prompt $\texttt{p}$. We first align the model representation, mapped by the capability encoder, into an embedded feature of LLM inputs. This is achieved using a single-layer MLP, which acts as a connector to adapt the dimensions. Consequently, we derive the aligned model capability vector:
\begin{equation}
    \mathbf{e}_{\texttt{c}^m} = \mathbf{W} \cdot \boldsymbol{\psi} \left( \texttt{c}^m \right)\,,
\end{equation}
where $\mathbf{e}_{\texttt{c}^m}$ is combined with the input embeddings of $\mathbf{x}_i$ and $\texttt{p}$ to form the capability instruction $\mathbf{z}_{i}$, \textit{i.e.},
\begin{equation}
    \mathbf{z}_{i} = [\mathbf{e}_{\texttt{c}^m}, \mathbf{e}_{\mathbf{x}_i}, \mathbf{e}_{\texttt{p}}] = [\mathbf{e}_1, \mathbf{e}_2, \ldots, \mathbf{e}_s]\,,
\end{equation}
where $s$ denotes the length of the concatenated capability instruction sequence. Our alignment module operates at a natural language level, allowing for a streamlined design.  In the following Section~\ref{sec:experiments}, we also explore alternative approaches, including training without the alignment module.

\subsection{Tuning Recipe}

\textbf{Forward Process of the Prediction Score}: As shown in Figure~\ref{fig:capability_instruction}, the query prompt $\texttt{p}$ in the capability instruction includes keywords related to positive terms that the model excels at. For example, ``Yes'' serves as the key response in the prompt ``predict whether the model can handle test instruction by indicating `Yes' or `No'.'' In this context, the model routing prediction score is:
\begin{equation}
    \operatorname{Pr}\left( \text{`Yes'} \mid \mathbf{z}_i \right) = \prod_{t=1}^{s} \mathbf{1}_{\left(|\texttt{c}^m|, s\right]} \cdot \boldsymbol{\varphi} \left( \mathbf{e}_t \mid \left[ \mathbf{e}_1, \cdots, \mathbf{e}_{t-1} \right] \right)\,,
\end{equation}
where we omit the input embedding layer for the LLM $\boldsymbol{\varphi}$.

\input{tables/main_results}

\textbf{Positive and Negative Instructions for Training}:
We apply Homogeneous In-Batch Negative Sampling~\cite{DBLP:conf/emnlp/KarpukhinOMLWEC20,DBLP:journals/corr/abs-2310-07554} for each capability representation $\texttt{c}^m$ with its well-performed and poorly-performed instructions to enhance the discriminative during training.
Typically, a $k$-shot training batch $\mathbf{Z} = \left\{ \mathbf{z}_i \right\}_{i=0}^{k}$ contains $1$ positive instruction and $k-1$ negative ones.

\textbf{Loss Design}:
We denote the position of the positive instruction in the training batch $\mathbf{Z}$ as $y_{\text{pos}}$, and the remaining ones are $k-1$ negative instructions.
Our objective is to enhance the prediction score for the positive ones as the candidate performs better on this instruction.
We employ the cross-entropy loss to optimize this in one batch $\mathbf{Z}$:
\begin{equation}
    \mathcal{L}_{\text{CE}} = \mathbb{E}_{\mathbf{Z} \in \mathcal{D}_{\text{test}}} \left[ - \log \operatorname{Pr} \left(\boldsymbol{h}_{\boldsymbol{\varphi}, \text{`Yes'}}\left(\mathbf{Z}\right) = y_{\text{pos}} \, | \, \mathbf{Z} \right) \right]\,,
\end{equation}
where $\boldsymbol{h}_{\boldsymbol{\varphi}, \text{`Yes'}}\left(\mathbf{Z}\right) = \arg\max_{\mathbf{z}_i \in \mathbf{Z}} \, \operatorname{Pr} \left(\text{`Yes'} \, | \, \mathbf{z}_i \right)$ is the LLM $\boldsymbol{\varphi}$ to identify which instruction $\mathbf{z}_i \in \mathbf{Z}$ can be done well (positive) and which cannot (negative).

\textbf{Learning Strategy}: The model representation is derived from the capability distribution on MMLU. Similarly, we develop both in-domain and out-of-domain learning environments for \textsc{Model-SAT}.
In the first stage, we collect in-domain positive and negative training instructions, primarily sourced from the same category as the MMLU dataset.
We only fine-tune the connector between the capability encoder $\boldsymbol{\psi}$ and the LLM $\boldsymbol{\varphi}$, establishing an initial capability-to-instruction mapping.
In the second stage, we fine-tune all model parameters.
We apply a larger learning rate on the encoder and connector to enhance capability alignment with instruction semantics.

\textbf{Data Refinement}: We further address noise in whether the candidate model can accurately perform the instructions, influencing whether a capability instruction is a positive or negative training sample.
For those difficult instructions that only a few models handle correctly, we implement a circle test by rotating the sequence of options to prevent lucky guesses.
Furthermore, we prioritize higher-ranked candidates in the training data by sampling with increased weight.

\subsection{Efficient Deployment}
\textsc{Model-SAT} provides the routing prediction for the candidate model applied to the target instruction.
These scores are generated by the same model, rendering them comparable.
In this paper, we propose an open-source and comprehensive model routing toolkit, \textsc{Model-SAT}.
This toolkit offers a viable solution for dynamic model routing within communities such as HuggingFace, harnessing the repository to boost performance on target tasks.

\textsc{Model-SAT} exhibits remarkable generalization capabilities for unseen data, which can be directly concatenated into the capability instruction.
Similarly, the incremental extension to new models proves highly efficient, requiring only inference on 50 core tasks for the model representation.
As later addressed in the experiments, \textsc{Model-SAT} exhibits zero-shot model routing abilities, facilitating the streamlined development of capability instructions in broader contexts.


\section{Experiments}
\label{sec:experiments}

This section begins by detailing the construction of training and test instructions in \textit{capability instructions tuning}. It then presents different zoo setups for testing and concludes with an analysis of results and ablation studies.

\subsection{Training and Test Instructions}

As mentioned earlier, the \textit{capability instruction} consists of model representation $\texttt{c}^m$, instructions $\mathbf{x}_i$ to assign, and performance inquiry prompts $\texttt{p}$.

\textbf{Candidate Model Representations} $\texttt{c}^m$ for candidate $m$: We introduced 66 open-source LLMs of varying scales. This includes 60 models under 10B, 15 ones between 10B and 20B, and 5 ones around 60B.
We sample 50 categories from the MMLU dataset, with 20 distinguishing instructions from each. Different candidate models share core tasks to ensure stability in capability demonstration.

\textbf{Instructions} $\mathbf{x}_i$ \textbf{Pending to Assign}: We consider more than 20 datasets that include areas such as language, analysis, understanding, and reasoning in general evaluations, as well as specialized fields like mathematics and medicine. For each dataset, we sample sets of positive and negative instructions where the model performed well or poorly, with sampling on stronger models assigned greater weight. Each dataset contains about 100 instructions on average.

\input{tables/compared_results}

\input{tables/mm_results}

\textbf{Performance Inquiry Prompts} $\texttt{p}$: We explore different approaches for the probability of model routing. For \textit{capability instructions}, we design the performance inquiry prompt, such as ``predict whether the model can handle test instruction by indicating `Yes' or `No'.'' In this context, a response of `Yes' signifies that the model is well-performed to the instruction. We also experiment with integrating a regression linear layer onto the next token embedding.

The capability instruction for the test $\mathbf{z}_i$ similarly consists of the model representation $\texttt{c}^m$, the target instruction $\mathbf{x}_i$ to be assigned, and the performance inquiry prompt $\texttt{p}$. To ensure test stability, we conduct a perturbation evaluation on model representation. Specifically, we randomly alter the ranking of the aptitude test results in capability representation twice and then calculate the average routing scores $\operatorname{Pr}\left( \text{`Yes'} \mid \mathbf{z}_i \right)$. The response on this instruction $\mathbf{x}_i$ is provided by the candidate model with the highest routing score.

\subsection{Benchmarks of LLM Routing}

\label{sec:benchmarks}
In this section, we outline benchmarks with various LLMs and their extension zoos, featuring detailed settings.

\begin{figure*}[t]
    \centering
    \vspace{-10pt}
    \includegraphics[width=0.97\textwidth]{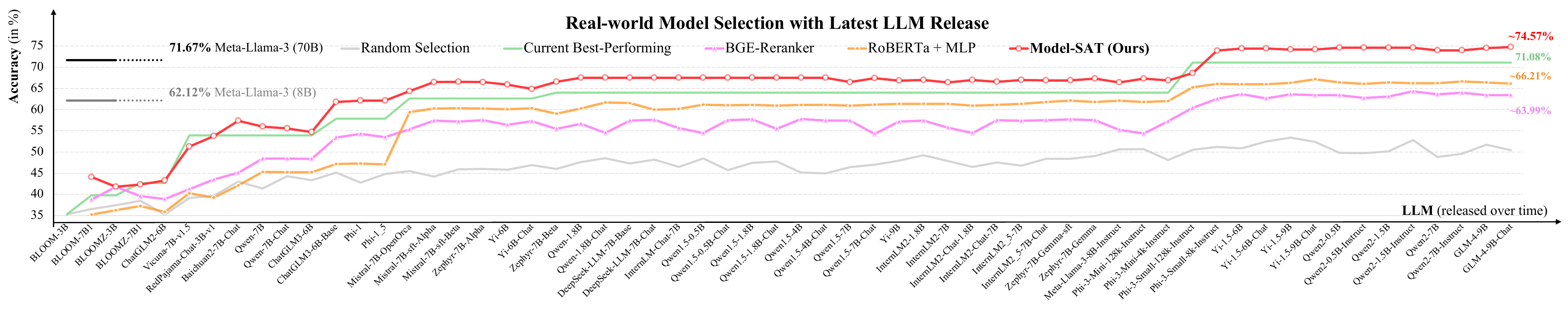}
    \caption{\textbf{Real-world Model Routing with Latest LLM Release} on ARC-Challenge. \textsc{Model-SAT} (in red) quickly generalizes to unseen LLMs without extra training, maintaining robust performance despite dynamically adding diverse LLMs.}
    \label{fig:exp_time}
    \vspace{-12pt}
\end{figure*}
\textbf{Smaller-Scale LLM do Better}: As demonstrated in the Table~\ref{tab:m_res}, the \textbf{smaller-scale zoo} contains InternLM2.5 (7.7B), Meta-Llama-3-Instruct (8.0B), Qwen2-Instruct (7.6B), GLM-4 (9.4B), and Phi-3-Small-128K (7.4B).
The \textbf{smaller-mixed zoo} includes the smaller-scale zoo and Phi-1 (1.3B), BLOOMZ (3B), and Zephyr-Alpha (7.2B).
These LLMs have fewer than 10B parameters and low deployment costs.
In Figure~\ref{fig:motivation}, we show that the union of correct responses can cover a set of instructions that only larger-scale ones can manage.

\textbf{General LLM Zoo Settings.} \textbf{1}) \textbf{Middle-Mixed and Larger-Mixed LLM Zoo}: The \textbf{middle-mixed zoo} includes the smaller-scale zoo and Phi-3-Medium-128K (14B), and Yi-1.5-Chat (34B). The \textbf{larger-mixed zoo} consists of the middle-mixed ones, Meta-Llama-3-Instruct (70B), Qwen2-Instruct (72B), and Mixtral-8x22B-Instruct-v0.1 (140B). The mixed zoo can validate the routing method across different capabilities.
\textbf{2}) \textbf{High-Performance LLM Zoo}: We select from larger-scale LLMs to boost performance further.
The model zoo contains only three models above with over 70B parameters.
\textbf{3}) \textbf{Multimodal LLM Zoo}: To verify the generality of capability instruction tuning, we construct a multimodal LLM zoo that includes MiniCPM-Llama3-V 2.5, Phi-3-Vision-128k-Instruct, and InternLM-XComposer2-VL-7B.

\textbf{Instructions for Model Routing Evaluation}:
The test capability instructions differ from the training ones of model routers and consist of seven evaluation datasets.
Datasets including MMLU~\cite{DBLP:conf/iclr/HendrycksBBZMSS21} (5-shot) and WinoGrande~\cite{DBLP:conf/aaai/SakaguchiBBC20} (5-shot) cover a broad range and are involved in the training part as the in-domain evaluation.
On the other hand, datasets such as ARC-Challenge~\cite{DBLP:journals/corr/abs-2102-03315} (25-shot), TruthfulQA (6-shot), and BoolQ~\cite{clark2019boolq} (1-shot) with MRPC (1-shot) and MNLI (1-shot) in GLUE~\cite{DBLP:conf/iclr/WangSMHLB19} benchmark focus on specific capabilities, serving as the unseen out-of-domain evaluation of the model routers.
We consider the evaluation datasets MMMU-VAL~\cite{mmlu}, AI2D-TEST~\cite{kembhavi2016ai2d}, and CCBench~\cite{liu2024mmbenchmultimodalmodelallaround} in the multimodal scenario.

\textbf{Real-world Model Routing with Unseen Datasets \& Latest LLMs}: \textbf{1}) In Table~\ref{tab:m_res}, \textit{In-Domain} and \textit{Out-of-Domain} indicate whether the dataset is included in the training set for LLM routing. \textbf{2}) In Figure~\ref{fig:exp_time}, We design a novel model routing setting that, with the \textbf{release of 60 LLMs}, we update the existing zoo after each new model with the top 5 historically best and the latest one.
With the continual increment of unseen LLMs, this dynamic environment tests whether methods can maintain a compelling performance.

\subsection{Toward Comprehensive and Effective Routing}
\label{sec:results}
\textbf{Performance Analysis in Various Model Zoos.} Table~\ref{tab:m_res} demonstrates that \textsc{Model-SAT} performs impressively across five comprehensive LLM Zoos.
(a) Smaller-Scale: routing of LLMs under 10B achieve performance comparable to the \textasciitilde70B LLMs. \textsc{Model-SAT}'s average score of 75.28\% closely matches Meta-Llama-3-Instruct-70B's 76.90\%, and outperforms it on the ARC-C and TruthfulQA benchmarks. Furthermore, \textsc{Model-SAT} selects the optimal model for each instruction, surpassing the best-performing models in the Smaller-Scale Zoo.
(b) Smaller-Mixed: We add three earlier released, weaker, and smaller LLMs. \textsc{Model-SAT} maintained stable performance, with a slight decrease of about 1\% compared to row (a), while performance on ARC-C, BoolQ, and MNLI benchmarks remained nearly identical.
(c) Middle-Mixed: Row (c) includes two medium-scale models (10B to 70B), resulting in improved performance for \textsc{Model-SAT} compared to row (a). Its average performance now closely matches that of the 70B model.
(d) Larger-Mixed: Incorporating three 70B models in row (d) showed that \textsc{Model-SAT} remains robust despite significant performance variances in the LLM zoo, with improvements of nearly 5\% on MMLU, BoolQ, and so on.
(e) High-Performance: Row (e) features a routing of only three 70B models, revealing that the capabilities of gigantic LLMs are further unleashed, achieving a state-of-the-art score of 85.64\% on MMLU and 73.63\% on the ARC-Challenge.

\textbf{Comparative Analysis of Routing Delay.} In Table~\ref{tab:m_res}, the selected model parameter-scale is denoted as xB. The overhead associated with the re-ranking method is related to the $M$ candidates of xB. Figure~\ref{fig:related_works} illustrates that the re-ranking requires obtaining all inference results from the zoo first, while \textsc{Model-SAT} processes model representations all at once and utilizes them throughout its lifetime. As the scale of models in the zoo grows, \textsc{Model-SAT}'s routing cost remains unaffected by inference ones, ensuring efficient model routing.

\textbf{Comparison Ranking-based Methods}: We evaluate re-ranking methods such as Cappy, BGE-Large, and GTE-Large. Although they have access to the outputs of each candidate, re-ranking often struggles to find optimal results, potentially because it primarily focuses on retrieving different semantics rather than optimizing performance across similar outputs.

\textbf{Detailed Comparative Analysis}: Furthermore, we explore various learning strategies within the capability instruction tuning framework. Using features extracted from RoBERTa, we train MLP (classification-based), $k$ Nearest Neighbors (clustering-based), and Random Forest (tree-based). We observed that, except for tree models that are suitable for handling capability representation as similar tabular data, other learning strategies fail to capture performance distribution mappings. Additionally, we analyze \textsc{Model-SAT} without the capability encoder. Since the model representation is expressed at the natural language level, Phi-3 in Table 2 can also learn some LLM-Instruction mappings, but its performance remains inferior to that of capability-encoder-based ones.

\textbf{Ablation Studies: Explore Generalization in Multimodal Scenarios.} Most multimodal LLMs (MLLMs) are derived from input-extending LLMs. Multimodal \textsc{Model-SAT} is built on the Wings~\cite{zhang2024wingslearningmultimodalllms} training architecture, integrating model representation embeddings with visual ones. It maintains strong performance in multimodal scenarios, achieving optimal average performance across MMMU-VAL, AI2D-TEST, and CCBench datasets.

\section{Conclusion}

This paper proposes a novel model routing paradigm called \textit{capability instruction tuning} with instruction-level model routing. It constructs a capability instruction comprising capabilities, instructions, and inquiry prompts to select the most suitable one. We present \textsc{Model-SAT}, featuring a capability encoder and lightweight LLM. It selects models without inference overhead of candidates and quickly adapts to new models. Model-SAT performs optimally across five proposed LLM routing settings and its multimodal extension.

%% file: tables/main_results.tex
\begin{table*}[t]
\small
\vspace{-10pt}
\centering
\setlength\tabcolsep{1.7pt}
\begin{tabular}{ p{2.6em} p{6.2em}<{\centering} p{4.8em}<{\centering} p{3.7em}<{\centering}  p{3.7em}<{\centering}  p{3.7em}<{\centering}  p{3.7em}<{\centering}  p{3.7em}<{\centering}  p{3.7em}<{\centering}  p{3.7em}<{\centering}  p{3.7em}<{\centering}}
\toprule
\multirow{2}{*}{\textbf{Method}} & & \multirow{2}{*}{\textbf{\#Params}} & \multicolumn{2}{c}{\textbf{In-Domain}} & \multicolumn{5}{c}{\textbf{Out-of-Domain}} & \multirow{2}{*}{\textbf{Mean}} \\
& & & MMLU & WinoG. & ARC-C & BoolQ & TruthfulQA & \, MRPC & MNLI & \\
\midrule
& & & & & \multicolumn{6}{l}{{\em \quad \quad \, Smaller-scale LLMs (<10B)}} \\
\noalign{\vskip 0.7ex}
\cline{6-9}
\noalign{\vskip 0.7ex}
\multicolumn{2}{l}{InternLM2.5{\footnotesize~\cite{cai2024internlm2}}} & 7.7B & 69.88 & 81.22 & 60.75 & 70.43 & 54.56 & 68.38 & 60.68 & 66.89 \\ 
\multicolumn{2}{l}{Meta-Llama-3$_{\text{\ Instruct}}${\footnotesize~\cite{touvron2023llama2}}} & 8.0B & 65.59 & 75.45 & 62.12 & 76.76 & 51.63 & 68.38 & 55.82 & 65.62 \\ 
\multicolumn{2}{l}{Qwen2$_{\text{\ Instruct}}${\footnotesize~\cite{yang2024qwen2technicalreport}}} & 7.6B & 69.13 & 74.11 & 61.43 & 82.57 & 55.49 & 78.92 & 54.96 & \underline{68.19} \\ 
\multicolumn{2}{l}{GLM-4{\footnotesize~\cite{DBLP:journals/corr/abs-2406-12793}}} & 9.4B & 69.28 & 80.82 & 66.13 & 84.77 & 59.32 & 78.92 & 40.73 & 68.65 \\
\multicolumn{2}{l}{Phi-3$_{\text{\ Small-128K}}${\footnotesize~\cite{DBLP:journals/corr/abs-2404-14219}}} & 7.4B & 75.90 & 77.11 & 71.08 & 86.70 & 64.62 & 75.98 & 46.82 & 71.38 \\
\multicolumn{2}{l}{Best-Performing among the five above} & xB & 75.90 & 81.22 & 71.08 & 86.70 & 64.62 & 78.92 & 60.68 & 74.16\\
\noalign{\vskip 0.7ex}
& & & \multicolumn{8}{l}{\quad \quad \quad \quad \quad { \textbf{LLM Selection} on \textit{Smaller-scale} LLMs (5 models)}} \\
\noalign{\vskip 0.7ex}
\cline{6-9}
\noalign{\vskip 0.7ex}
\multicolumn{2}{l}{{Random Selection}} & xB & 70.11 & 77.66 & 64.59 & 79.94 & 57.58 & 72.79 & 51.54 & 67.74\\
\rowcolor{mygreyr}
\multicolumn{2}{l}
{Cappy{\footnotesize~\cite{DBLP:journals/corr/abs-2311-06720}}} & $M\cdot$xB & 69.53 & 78.06 & 63.99 & 81.10 & 57.77 & 74.75 & 53.17 & 68.34 \\
\rowcolor{mygreyr2}
\multicolumn{2}{l}{BGE$_{\text{\ Large}}${\footnotesize~\cite{bge_embedding}}} & (0.3+$M$)$\cdot$xB & 71.68 & 78.53 & 66.38 & 82.48 & 61.44 & 73.28 & 55.39 & 69.88 \\
\rowcolor{mygreyr4}
\multicolumn{2}{l}{GTE$_{\text{\ Large}}${\footnotesize~\cite{zhang2024mgte}}} & (1.8+$M$)$\cdot$xB & 72.02 & 79.32 & 68.09 & 83.73 & 59.61 & 75.74 & 56.16 & 70.67\\
\rowcolor{myred}
\multicolumn{2}{l}{\textsc{Model-SAT} \textbf{(Ours)}} & $M\cdot$4.3B+xB & 79.86 & 82.24 & 72.53 & 86.73 & 65.12 & 79.66 & 60.83 & 75.28\\
\bottomrule
\noalign{\vskip 0.7ex}
\noalign{\vskip 0.7ex}
& & & & & \multicolumn{6}{l}{{\em \quad \quad Larger-Scale LLMs (10B\textasciitilde50B)}} \\
\noalign{\vskip 0.7ex}
\cline{6-9}
\noalign{\vskip 0.7ex}
\multicolumn{2}{l}{Phi-3$_{\text{\ Medium-128K}}${\footnotesize~\cite{DBLP:journals/corr/abs-2404-14219}}} & 14B & 76.63 & 74.35 & 66.49 & 86.30 & 54.54 & 78.92 & 59.42 & 70.95\\
\multicolumn{2}{l}{Yi-1.5$_{\text{\ Chat}}${\footnotesize~\cite{DBLP:journals/corr/abs-2403-04652}}} & 34B & 77.15 & 81.47 & 70.62 & 87.84 & 62.02 & 80.88 & 61.56 & 74.50 \\
\multicolumn{2}{l}{Meta-Llama-3$_{\text{\ Instruct}}${\footnotesize~\cite{touvron2023llama2}}} & 70B & 79.89 & 82.62 & 71.67 & 93.61 & 61.83 & 83.58 & 65.07 &76.90 \\ 
\multicolumn{2}{l}{Qwen2$_{\text{\ Instruct}}${\footnotesize~\cite{yang2024qwen2technicalreport}}} & 72B & \underline{83.79} & 84.41 & 68.62 & \underline{94.90} & 54.85 & \underline{84.31} & 66.95 & 76.83\\ 
\multicolumn{2}{l}{Mixtral-8x22B$_{\text{\ Instruct-v0.1}}${\footnotesize~\cite{jiang2024mixtral}}} & 140B & 77.63 & \underline{85.25} & \underline{72.68} & 92.71 & \underline{68.19} & 81.13 & \underline{67.70} & \underline{77.90}\\
\midrule
\noalign{\vskip 0.7ex}
& & & & \multicolumn{6}{l}{\quad \, \textbf{Capability Instruction Tuning} w/o Inference Overhead} \\
\noalign{\vskip 0.7ex}
\cline{5-10}
\noalign{\vskip 0.7ex}
\rowcolor{myred}
& \multicolumn{1}{l}{{\scriptsize Smaller-Scale LLM Zoo}} &  & 79.86 & 82.24 & 72.53 & 86.73 & 65.12 & 79.66 & 60.83 & 75.28\\
\rowcolor{myred1}
& \multicolumn{1}{l}{{Smaller-Mixed LLM Zoo}} &  & 78.60 & 82.08 & 72.01 & 86.48 & 64.50 & 78.19 & 60.72 & 74.65\\
\rowcolor{myred2}
\textbf{Ours} & \multicolumn{1}{l}{{Middle-Mixed LLM Zoo}} & $M\cdot$4.3B+xB & 79.97 & 83.03 & 72.69 & 87.80 & 64.87 & 83.82 & 61.80 & 76.28\\
\rowcolor{myred3}
& \multicolumn{1}{l}{{Larger-Mixed LLM Zoo}} & & 84.16 & 86.27 & 73.21 & 93.94 & 69.16 & 85.54 & 67.92 & 80.03\\
\rowcolor{myred4}
& \multicolumn{1}{l}{\textbf{High-Performance LLM Zoo}} & & \textbf{85.64} & \textbf{87.85} & \textbf{73.63} & \textbf{95.02} & \textbf{69.40} & \textbf{88.24} & \textbf{68.39} & \textbf{81.17}\\
\bottomrule
\end{tabular}
\caption{\textbf{A Comprehensive Performance Evaluation}: Covering smaller-scale, high-performance giant LLMs, and a mixed LLM zoo of small, medium, and large levels. Model-SAT performs instruction-level model selection, consistently maintaining efficient and precise results that outperform the optimal one in the LLM zoo. Bold is the best, and underlined is the second-best.}
\vspace{-10pt}
\label{tab:m_res}
\end{table*}

%% file: tables/compared_results.tex
\begin{table}[t]
\small
\vspace{-6pt}
\label{tab:compared_results}
\centering
\setlength\tabcolsep{1.7pt}
\begin{tabular}{ p{2.6em} p{6.2em}<{\centering} p{3.5em}<{\centering} p{3.5em}<{\centering}  p{3.5em}<{\centering} p{3.5em}<{\centering}}
\toprule
\multicolumn{2}{l}{\textbf{Method}} & {\scriptsize \textbf{MMLU}} & {\scriptsize \textbf{ARC-C}} & {\scriptsize \textbf{TruthfulQA}} & {\scriptsize \textbf{Mean}} \\
\midrule
\multicolumn{2}{l}{Random Selection} & 70.11 & 64.59 & 57.58 & 64.09 \\
\multicolumn{2}{l}{Best-Perfoming} & \underline{75.90} & \underline{71.08} & \underline{64.62} & \underline{70.53} \\
\multicolumn{2}{l}{RoBERTa + {\scriptsize MLP}} & 71.25 & 64.33 & 59.12 & 64.90 \\
\multicolumn{2}{l}{\,\,\,\,+ {\scriptsize $k$ Nearest Neighbors}} & 70.02 & 65.02 & 58.51 & 64.52 \\
\multicolumn{2}{l}{\,\,\,\,+ {\scriptsize Random Forest}} & 73.75 & 68.05 & 61.44 & 67.75 \\
\multicolumn{2}{l}{Phi-3$_{\text{\ Mini-128K}}$} & 74.71 & 70.58 & 61.70 & 68.83 \\
\multicolumn{2}{l}{\textsc{Model-SAT}} & \textbf{79.86} & \textbf{72.53} & \textbf{65.12} & \textbf{72.50} \\
\bottomrule
\end{tabular}
\caption{\textbf{Performance Comparisons of Other Learning Strategies for Capability Instructions}. The capability encoder of Model-SAT learns the mapping of performance to semantics, demonstrating strong model selection abilities.}
\vspace{-14pt}
\end{table}

%% file: tables/mm_results.tex
\begin{table}[t]
\small
\vspace{-6pt}
\label{tab:mm_results}
\centering
\setlength\tabcolsep{1.7pt}
\begin{tabular}{ p{5.3em} p{3.6em}<{\centering} p{3.6em}<{\centering} p{3.6em}<{\centering}  p{3.6em}<{\centering} p{4.5em}<{\centering}}
\toprule
\multirow{2}{*}{\textbf{Dataset}} & \raisebox{-0.4ex}{\scriptsize{\textbf{Phi-3}}} & \raisebox{-0.4ex}{\scriptsize{\textbf{InternLM}}} & \raisebox{-0.4ex}{\scriptsize{\textbf{MiniCPM}}} & \raisebox{-0.4ex}{\scriptsize{\textbf{Random}}} & \raisebox{-0.4ex}{\scriptsize{\textbf{\textsc{Model-SAT}}}} \\
& \raisebox{0.4ex}{\tiny{\textbf{Vision}}} & \raisebox{0.4ex}{\tiny{\textbf{XC2-VL}}} & \raisebox{0.4ex}{\tiny{\textbf{Llama3-V}}} & \raisebox{0.4ex}{\tiny{\textbf{Selection}}} & \raisebox{0.4ex}{\scriptsize{\textbf{(Ours)}}} \\
\midrule
MMMU{\scriptsize \,VAL} & 41.86 & 41.06 & \underline{43.10} & 42.19 & \textbf{43.21} \\
AI2D{\scriptsize \,TEST} & 78.40 & \underline{79.44} & 77.33 & 78.24 & \textbf{80.38} \\
CCBench & 37.60 & 56.62 & \textbf{57.16} & 50.49 & \underline{56.96} \\
\bottomrule
\end{tabular}
\caption{\textbf{Performance Comparisons in Selection of Multimodal LLM}. Model-SAT maintains excellent average performance on the above three popular evaluation benchmarks.}
\vspace{-14pt}
\end{table}